\documentclass{article}
\usepackage{ijcai23}

\usepackage{times}
\usepackage{soul}
\usepackage{url}
\usepackage[hidelinks]{hyperref}
\usepackage[utf8]{inputenc}
\usepackage[small]{caption}
\usepackage{graphicx}
\usepackage{amsmath}
\usepackage{amsthm}
\usepackage{amssymb}
\usepackage{booktabs}
\usepackage{algorithm}
\usepackage{cleveref}
\usepackage{caption}
\usepackage{subcaption}
\usepackage{multirow}
\usepackage{multicol}
\usepackage{adjustbox}
\usepackage{amsmath}
\usepackage{color}
\usepackage{breqn}
\usepackage{dsfont}
\usepackage{algpseudocode}
\usepackage{tabto}

\usepackage{float} 

\title{Distilling Universal and Joint Knowledge for Cross-Domain Model Compression on Time Series Data}

\author{
Qing Xu$^{1,3}$
\and
Min Wu$^1$\and
Xiaoli Li$^{1,2}$\And
Kezhi Mao$^3$ \and
Zhenghua Chen$^{1,2}$\footnote{Corresponding author}
\affiliations
$^1$Institute for Infocomm Research, A*STAR, Singapore\\
$^2$Centre for Frontier AI Research, A*STAR, Singapore\\
$^3$Nanyang Technological University, Singapore
\emails
\{Xu\_Qing, wumin, xlli\}@i2r.a-star.edu.sg,
EKZMao@ntu.edu.sg,
chen0832@e.ntu.edu.sg
}

\begin{document}

\maketitle

\begin{abstract}
  
For many real-world time series tasks, the computational complexity of prevalent deep leaning models often hinders the deployment on resource-limited environments (\textit{e.g.,} smartphones). Moreover, due to the inevitable domain shift between model training (source) and deploying (target) stages, compressing those deep models under cross-domain scenarios becomes more challenging. Although some of existing works have already explored cross-domain knowledge distillation for model compression, they are either biased to source data or heavily tangled between source and target data. To this end, we design a novel end-to-end framework called \textbf{UN}iversal and jo\textbf{I}nt \textbf{K}nowledge \textbf{D}istillation (\textbf{UNI-KD}) for cross-domain model compression. In particular, we propose to transfer both the universal feature-level knowledge across source and target domains and the joint logit-level knowledge shared by both domains from the teacher to the student model via an adversarial learning scheme. More specifically, a feature-domain discriminator is employed to align teacher's and student's representations for universal knowledge transfer. A data-domain discriminator is utilized to prioritize the domain-shared samples for joint knowledge transfer. Extensive experimental results on four time series datasets demonstrate the superiority of our proposed method over state-of-the-art (SOTA) benchmarks. The source code is available at \href{https://github.com/ijcai2023/UNI\_KD} {https://github.com/ijcai2023/UNI\_KD}.



\end{abstract}

\section{Introduction}
\label{sec:introduction}
Deep learning (DL) models, particularly convolutional neural networks (CNNs), have achieved remarkable successes in various time series tasks, such as human activity recognition (HAR) \cite{wang2019deep}, sleep stages classification \cite{eldele2021attention} and fault diagnosis \cite{zhao2019deep}. These advanced DL models are often over-parameterized for better generalization on unseen data \cite{chen2017learning}. However, deploying those models on a resource-limited environment (\textit{e.g.,} smartphones and robots) is a common requirement for many real-world applications. The contradiction between model performance and complexity leads to the exploration of various model compression techniques, such as network pruning and quantization \cite{liang2021pruning}, network architecture search (NAS) \cite{elsken2019neural} and knowledge distillation (KD) \cite{hinton2015distilling}. Among them, KD has demonstrated its superior effectiveness and flexibility on enhancing the performance of a compact model (\textit{i.e.}, Student) via transferring the knowledge from a cumbersome model (\textit{i.e.}, Teacher). Another well-known problem in many time series tasks is the considerable domain shift between model development and deployment stages. For instance, due to the difference between subject's genders, ages or data collection sensors, a model trained on one subject (\textit{i.e.}, source domain) might perform poorly on another subject (\textit{i.e.}, target domain). Such domain disparity makes cross-domain model compression even more challenging. 


Some recent works have already attempted to explore the benefits of applying unsupervised domain adaption (UDA) techniques during compressing cumbersome DL models by knowledge distillation. However, there are some drawbacks in these approaches. For instance, joint training of a teacher with UDA and student with KD would result in unstable loss convergence \cite{granger2020joint}, while the knowledge from teachers trained on source domain only \cite{yang2020mobileda,ryu2022knowledge} is biased and limited. For cross-domain knowledge distillation, a proper teacher should possess the knowledge of both domains. In particular, the generalized knowledge (namely \textit{Universal Knowledge}) across both domains is more critical in improving student's generalization capability on target domain. However, the aforementioned methods coarsely align teacher's and student's predictions, but neglect to disentangle the domain-shared knowledge (namely \textit{Joint Knowledge}). Due to the existence of domain shift, introducing source-specific knowledge would result in poor adaptation performance. 

\begin{figure}[t]
    \centering
    \includegraphics[width=0.5\textwidth]{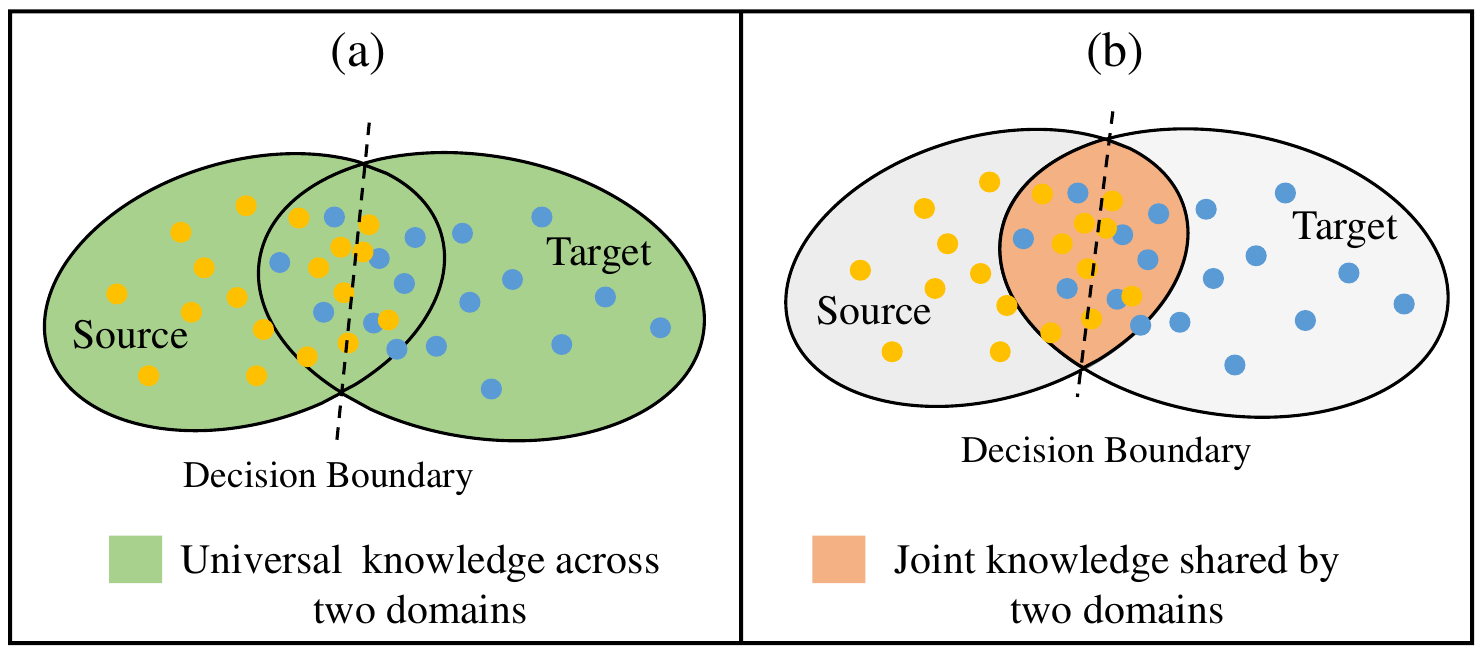}
    \caption{Illustration of universal and joint knowledge.}
    \label{fig:figure-domain-knowledge}
\end{figure}

Fig. \ref{fig:figure-domain-knowledge} presents an example of our proposed universal and joint knowledge under cross-domain scenario. On the one hand, the universal knowledge across source and target domains as shown in Fig. \ref{fig:figure-domain-knowledge}(a) is important to improve the generalization capability for the student. On the other hand, the inevitable domain shift makes the distributions of source and target domains overlapped. Suppose that there exists a data-domain discriminator to correctly classify the samples into source or target domain. As depicted in Fig. \ref{fig:figure-domain-knowledge}(b), if some samples lie around its decision boundary, then these samples most likely possess some domain-shared information (\textit{i.e.}, joint knowledge) which makes the discriminator incapable of correctly identifying their data domain (\textit{i.e.}, source or target). Meanwhile, those samples which can be very confidently identified by the data-domain discriminator tend to possess domain-specific knowledge. Equally treating all samples like conventional KD approaches would be adverse to diminishing domain disparity, leading to poor generalization on target data. It is thus highly motivated to pay more attentions on samples with joint knowledge than samples with domain-specific knowledge for cross-domain knowledge distillation. 

In this paper, we propose an innovative end-to-end model compression framework to improve student's generalization capability under the cross-domain scenarios. Specifically, we design a feature-domain discriminator to align teacher's and student's feature representations for effectively distilling the universal knowledge. Meanwhile, a data-domain discriminator is developed to prioritize the samples with joint knowledge across two domains. It assists to disentangle teacher's logits by paying more attentions on the samples with joint knowledge. Via an adversarial learning scheme, teacher's universal and joint knowledge can be effectively transferred to the compact student. Our main contributions are summarized as follows.
\begin{itemize}
    \item A novel approach named universal and joint knowledge distillation (UNI-KD) approach is proposed to transfer teacher's universal and joint knowledge, which is an end-to-end framework for cross-domain model compression. Two discriminators (\textit{i.e.}, feature-domain and data-domain discriminators) with an adversarial learning paradigm are designed to distill above two knowledge on feature-level and logit-level, respectively.

    \item We propose to disentangle teacher's logits with a data-domain discriminator by prioritizing the samples with joint knowledge across source and target domains. The joint knowledge could further boost the generalization ability of the compact student on target domain. 

    \item Extensive experiments are conducted on four real-world datasets across three different time series classification tasks and the results demonstrate the superiority of our approach over other SOTA benchmarks. 
\end{itemize}

\begin{figure*}[ht]
    \centering
    \includegraphics[width=0.9\textwidth]{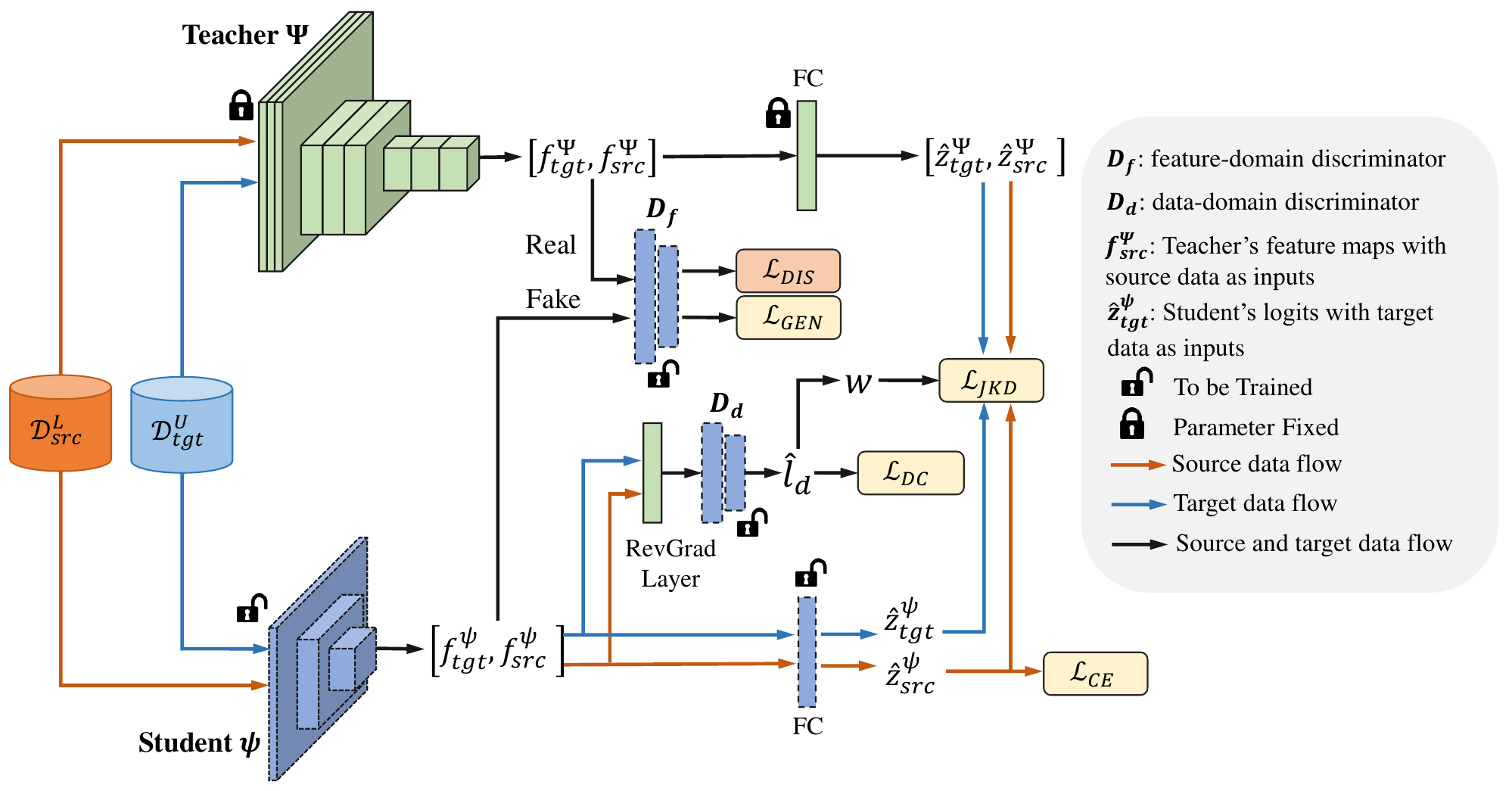}
    \caption{Illustration of proposed UNI-KD. A feature-domain discriminator $D_f$ is employed to identify whether the input feature maps are from teacher $\Psi$ or student $\psi$. It contributes to transferring the universal knowledge on feature level. A data-domain discriminator $D_d$ is leveraged to identify whether the input feature maps are from source or target domain. The output of $D_d$ is utilized to prioritize the samples for joint knowledge distillation on logits level. $D_f$ is adversarially trained against $D_d$ and student $\psi$.}
    \label{fig:figure-framework}
\end{figure*}

\section{Related Work}
\label {sec:related_work}

Knowledge distillation, as one of the most popular model compression techniques, has been widely explored in many applications. Originally, the knowledge from a complex teacher model is formulated as the logits soften by a temperature factor in  \cite{hinton2015distilling}. Then, researchers extend the knowledge to the feature maps as they contain more low-level information than logits. Several works try to minimize the discrepancy between teacher's and student's feature representations via explicitly defining distance metrics, such as L2 \cite{romero2014fitnets}, attention maps \cite{zagoruyko2016paying}, probability distributions \cite{passalis2018learning} and inter-channel correlation matrices \cite{liu2021exploring}. On the contrary, other researchers exploit the adversarial learning scheme which implicitly forces the student to generate similar feature maps as the teacher \cite{gao2019adversarial,xu2021kdnet,xu2022contrastive}. However, these approaches cannot be directly applied to cross-domain scenarios as they do not consider the domain shift during the compression. 


To tackle the domain shift issue, various UDA approaches have been proposed. Generally, these techniques can be categorized into two types, namely discrepancy-based and adversarial learning-based. The former ones intend to minimize some statistical distribution measurements between source and target domains, \textit{e.g.,} maximum mean discrepancy (MMD) \cite{tzeng2014deep}, the second-order statistics \cite{rahman2020minimum,sun2016deep} or higher-order moment matching (HoMM) \cite{chen2020homm}. Whereas the adversarial learning-based ones attempt to learn domain-invariant representations via a domain discriminator \cite{ganin2016domain,long2018conditional,shu2018dirt,wilson2020multi}. Although above mentioned UDA approaches have been successfully applied to many research areas, they seldom consider model complexity issue during domain adaptation, which is more practical for many time series tasks.  

Recently, there are some attempts to jointly address model complexity and domain shift problems by integrating UDA techniques with KD for cross-domain model compression. In \cite{granger2020joint}, a framework was proposed to employ the MMD to learn domain-invariant representations for teacher and progressively distill the knowledge to the student on both source and target data. However, their approach would lead to difficulty on student's convergence. MobileDA \cite{yang2020mobileda} performed the distillation on target domain with the knowledge from a source-only teacher. It leveraged the correlation alignment (CORAL) loss to learn domain-invariant representations for student. Similarly, a framework was proposed to perform adversarial learning and distillation on target domain with the knowledge from a source-only teacher in \cite{ryu2022knowledge}. The teachers in \cite{yang2020mobileda} and \cite{ryu2022knowledge} are trained on source data only and the knowledge from such teachers is very biased and limited. Unlike them, our method employs a teacher trained on labeled source domain and unlabeled target domain, and we distill not only the universal feature-level knowledge across both domains but also the joint logit-level knowledge shared by both domains via an adversarial learning scheme. 


\section{Methods}
\label{sec:methodology}

\subsection{Problem Definition}
For the cross-domain model compression scenario, we first assume that a proper teacher $\Psi$ is pre-trained on source and target domain data with SOTA UDA methods (\textit{e.g.,} DANN \cite{ganin2016domain}). Our objective is to improve the generalization capability of the compact student $\psi$ on target domain data. Same as other UDA works, we assume that data come from two domains: \textit{source} and \textit{target}. Data from source domain are labeled, $\mathcal{D}_{src}^{L} = \{x_{src}^i, y_{src}^i\}^{N_{src}}_{i=1}$, and data from target domain are collected from a new environment without labels, $\mathcal{D}_{tgt}^{U} = \{x_{tgt}^i\}^{N_{tgt}}_{i=1}$. Here, $N_{src}$ and $N_{tgt}$ refer to the number of samples in source and target domains, respectively. Let $\mathcal{P}(X_{src})$ and $\mathcal{Q}(X_{tgt})$ be the marginal distributions of two domains. UDA problems assume that $\mathcal{P}(X_{src}) \neq \mathcal{Q}(X_{tgt})$ but $\mathcal{P}(Y_{src}|X_{src}) = \mathcal{Q}(Y_{tgt}|X_{tgt})$, indicating that source and target domains have different data distributions but share a same label space. 

Our proposed UNI-KD is depicted as Fig. \ref{fig:figure-framework}. In order to effectively compress the model for the cross-domain scenario, we formulate teacher's knowledge into two categories: feature-level universal knowledge across two domains and logit-level joint knowledge shared by two domains. These two types of knowledge are complementary to each other. We introduce two discriminators with the adversarial learning scheme to efficiently transfer these two types of knowledge. 


\subsection{Universal Knowledge Distillation}
For cross-domain KD, we define the generalized knowledge across both domains as the universal knowledge. Such universal knowledge contains the fundamental characteristics existing in both source and target domains. In order to align teacher's and student's feature representations, adversarial learning scheme is utilized as it is capable of enhancing student's robustness \cite{maroto2022benefits}. Previous works also demonstrate that it could improve student's generalization capability on unseen data \cite{xu2022contrastive,chung2020feature}. Motivated by this, we first design a feature-domain discriminator $D_f$ for transferring the universal feature-level knowledge. $D_f$ is a binary classification network to identify the source of input feature, \textit{i.e.,} whether the input features $[f_{src}, f_{tgt}]$ come from $\Psi$ or $\psi$. 

We train $D_f$ and student $\psi$ in an adversarial manner. To be specific, in the first step, we fix student $\psi$ and train $D_f$ via loss $\mathcal{L}_{DIS}$ as Eq. (\ref{eq:equation-loss-dis}) shows. A batch of `real' samples (feature maps from teacher with source and target domain samples as input) is forwarded through $D_f$ to calculate the loss $\log(D_f(f_x^{\Psi}))$. The gradients are calculated with back propagation. Then a batch of `fake' samples (feature maps from student with same inputs) is forwarded through $D_f$ to calculate loss $\log(1-D_f(f_x^{\psi}))$. The gradients are accumulated to previous gradients from `real' samples. At last, minimizing $\mathcal{L}_{DIS}$ maximizes the probability of correctly classifying the input features as `real' (from teacher) or `fake' (from student). The second step is to fix the $D_f$ and train the student to generate similar feature maps as teacher. By minimizing $\mathcal{L}_{GEN}$ in Eq. (\ref{eq:equation-loss-gen}), the discriminator $D_f$ is expected to be incapable of telling whether the features are from $\Psi$ or $\psi$. Alternately applying above two steps over all the training samples forces the student to learn similar feature maps as the teacher.

\begin{equation}
\mathcal{L}_{DIS}= -\mathbb{E}_{x \sim (\mathcal{D}_{src},\mathcal{D}_{tgt})}[\log(D_f(f_x^{\Psi})) + \log(1-D_f(f_x^{\psi}))],
\label{eq:equation-loss-dis}
\end{equation}

\begin{equation}
\mathcal{L}_{GEN}= \mathbb{E}_{x \sim (\mathcal{D}_{src},\mathcal{D}_{tgt})}[\log(1-D_f(f_x^{\psi}))].
\label{eq:equation-loss-gen}
\end{equation}

However, there are some challenges for the above adversarial learning scheme. Firstly, it can only transfer the universal knowledge but neglect the domain disparity between source and target domains, resulting in poor generalization on target domain. Secondly, the optimization of student $\psi$ heavily relies on the accuracy of $D_f$ and the student would be difficult to converge especially in the early training stage. Thus, we introduce three additional losses in the second step of adversarial learning scheme to cope with the above issues in the following section. 

\subsection{Joint Knowledge Distillation}
Logits from teacher contain more information compared to one-hot labels and thus could be utilized as `dark' knowledge for distillation \cite{hinton2015distilling}. However, we empirically found that simply combining conventional logits knowledge with feature distillation might lead to performance degradation. Due to the existence of domain shift, knowledge from teacher can be divided into domain-joint knowledge shared by two domains and domain-specific knowledge only existing in a particular domain. Since we feed both source and target domain samples to teacher and student, roughly minimizing their logits distributions would transfer both knowledge, leading to poor transferring performance. 


Therefore, we intend to transfer the domain-joint knowledge but not domain-specific knowledge to the student. To achieve this, we utilize a data-domain discriminator $D_d$ whose output is a binary probability vector $\hat{l}_d = [p_{c=0}, p_{c=1}]$. The element in above vector represents the probability of the inputs belonging to source domain ($c=0$) or target domain ($c=1$). We argue that the samples lying around the distribution boundary of source and target domains in the feature space are more generic than those samples which can be classified with high confidence. In other words, if the data-domain discriminator $D_d$ cannot distinguish certain sample in the feature space, this sample most likely belongs to $\mathcal{P}(X_{src}) \bigcap \mathcal{Q}(X_{tgt})$ and possesses more domain-joint knowledge than others. Mathematically, $p_{c=0}$ and $p_{c=1}$ should be close to each other for these samples. Thus, we can utilize $\hat{l}_d$ to disentangle teacher's logits and let the student pay more attentions on those low-confidence samples during logits distillation. Specifically, for each sample $i$, we assign a different weight $w_i$ to adjust its contribution for logits-level knowledge distillation in Eq. (\ref{eq:equation-loss-weight-w}).

\begin{equation}
w_i= 1 - |p_{c=0}^i - p_{c=1}^i|. 
\label{eq:equation-loss-weight-w}
\end{equation}

Then, the loss $\mathcal{L}_{JKD}$ for joint knowledge distillation can be formulated as Eq. (\ref{eq:equation-loss-jkd}) where KL represents the Kullback-Lerbler divergence and 
$\textbf{\textit{q}}^s$, $\textbf{\textit{q}}^t \in \mathbb{R}^C$ are the predictions of student and teacher, respectively. $C$ is the number of classes. Each element $q_j$ in $\textbf{\textit{q}}^s$ or $\textbf{\textit{q}}^t$ is the probability of input sample belonging to the $j^{th}$ class and $j \in \{1,...,C\}$. $q_j$ is a function of temperature factor $\tau$ used for smoothing the distribution and can be calculated via Eq. (\ref{eq:equation-soft-targets}), and $z_j$ represents model outputs (\textit{i.e.,} logits) before the softmax layer. 


\begin{equation}
\mathcal{L}_{JKD}= \tau ^2 *\frac{1}{N} \sum_{i=0}^{N} w_i * KL (~\textbf{\textit{q}}^s||~ \textit{\textbf{q}}^t),
\label{eq:equation-loss-jkd}
\end{equation}

\begin{equation}
    q_j = \frac{exp(z_j/\tau)}{\sum_k{exp(z_k/\tau)}}.
\label{eq:equation-soft-targets}
\end{equation}

It is foreseeable that the efficacy of disentangling domain-joint and domain-specific knowledge to a large extent depends on the accuracy of $D_d$. Therefore, we introduce a domain confusion loss $\mathcal{L}_{DC}$ to assist the training of $D_d$ in Eq. (\ref{eq:equation-loss-dc}), where $l_d$ and $\hat{l}_d$ are the data domain labels and predictions of data-domain discriminator $D_d$, respectively.


\begin{align}
 \mathcal{L}_{DC}  =  - \mathbb{E}_{x \sim (\mathcal{D}_{src},\mathcal{D}_{tgt})}& \left[l_d*\log \hat{l}_d + \right. \nonumber \\
 &\left. (1-l_d)*\log(1-\hat{l}_d)\right]. \label{eq:equation-loss-dc} 
\end{align}

Furthermore, overfitting might occur if we only utilize target domain samples to train the student. To avoid this, we also optimize the student on the source domain via a cross-entropy loss as Eq. (\ref{eq:equation-loss-ce}) shows.

\begin{equation}
\mathcal{L}_{CE}=  - \mathbb{E}_{(x_{src},y_{src}) \sim \mathcal{D}_{src}} \sum_c \left[\mathds{1}_[{y_{src}=c}]\log q^s_c \right].
\label{eq:equation-loss-ce}
\end{equation}
Then, the final loss for training the student $\psi$ in the second step of adversarial learning paradigm is formulated as below:

\begin{equation}
\mathcal{L}= \mathcal{L}_{GEN} + (1-\alpha)*\mathcal{L}_{DC} + \alpha*\mathcal{L}_{JKD}  + \beta * \mathcal{L}_{CE}.
\label{eq:equation-loss-total}
\end{equation}
Here, we introduce two hyperparameters: $\alpha$ to balance the importance between domain confusion loss and logits KD loss, and $\beta$ to adjust the contribution of $\mathcal{L}_{CE}$. For $\alpha$, intuitively we intend to place more weights on UDA to achieve a good data-domain discriminator first and then gradually increase the importance of JKD as the training process goes on. This strategy can assist to stabilize student's training progress at the early stage. At each training epoch $m$, the corresponding value of $\alpha$ can be calculated by Eq. (\ref{eq:equation-alpha}), where $M$ is the total number of epochs, $a$ and $b$ are the starting and end values of $\alpha$. In our experimental setting, we set $\alpha \in [0.1,0.9]$. The value of $\alpha$ is exponentially increased with training epochs. Meanwhile, we utilize the grid search approach to identify the optimal value of parameter $\beta$. 

\begin{equation}
\alpha = a* e ^{\frac{m}{M}\log {\frac{b}{a}}}.
\label{eq:equation-alpha}
\end{equation}
Algorithm \ref{alg:uda_kd} illustrates the details of our proposed UNI-KD for cross-domain model compression.

\begin{algorithm}
\caption{UNI-KD for cross-domain model compression}
\label{alg:uda_kd}
\begin{algorithmic}[1]
\Require A pre-trained teacher $\Psi$, a student model $\psi$, a feature-domain discriminator $D_f$, a data-domain discriminator $D_d$, source dataset $\mathcal{D}_{src}^{L}$ and target dataset $\mathcal{D}_{tgt}^{U}$.
\For {epoch $m \in [1, M]$}
    \State Randomly shuffle $\mathcal{D}_{src}^{L}$ and $\mathcal{D}_{tgt}^{U}$;
    \State Update $\alpha$ for current epoch via Eq. (\ref{eq:equation-alpha});
    \For{each mini-batch in $X_b \in (\mathcal{D}_{src}^{L}, \mathcal{D}_{tgt}^{U}$)}
        \State Fix the parameters of $\psi$ and $D_d$;
        \State Forward $X_b$ through $\Psi$ as `real' samples;
        \State Forward $X_b$ through $\psi$ as `fake' samples;
        \State Calculate $\mathcal{L}_{DIS}$ as Eq. (\ref{eq:equation-loss-dis}); 
        \State Optimize $D_f$ via minimizing $\mathcal{L}_{DIS}$;
        \State Fix the parameters of $D_f$;
        \State Calculate $\mathcal{L}_{GEN}$ as Eq. (\ref{eq:equation-loss-gen}) and $\mathcal{L}_{CE}$ as Eq. (\ref{eq:equation-loss-ce});
        \State Calculate $\mathcal{L}_{DC}$ as Eq. (\ref{eq:equation-loss-dc}) and $\mathcal{L}_{JKD}$ as Eq. (\ref{eq:equation-loss-jkd});
        \State Update $\psi$ and $D_d$ by minimizing $\mathcal{L}$ via Eq. (\ref{eq:equation-loss-total});
    \EndFor
\EndFor
\end{algorithmic}
\end{algorithm}

\section{Experiments}
\label{sec:experiments}

\subsection{Datasets}
We evaluate our method with four commonly-used time series classification datasets across three different real-world applications: human activity recognition, sleep stage classification and fault diagnosis.

\paragraph{UCI HAR} \cite{anguita2013public}: A smartphone is fixed on the waist of 30 experimental subjects and each subject is requested to perform six activities, \textit{i.e.,} walking, walking upstairs, walking downstairs, standing, laying and sitting. The measurements from accelerometer and gyroscope are recorded for identifying each activity. Due to the variability between different subjects, we consider each subject as an independent domain and randomly select five cross-domain scenarios same as \cite{ragab2022adatime} for evaluation.

\paragraph{HHAR} \cite{stisen2015smart}: In this dataset, each subject conducts six activities, \textit{i.e.,} biking, sitting, standing, walking, walking upstairs and downstairs. Since different brands of smartphones and smart watches are leveraged for data collection, this dataset is considered more challenging than \textbf{UCI HAR} in terms of domain shift. We follow \cite{liu2021adversarial} and select five cross-domain scenarios for evaluation.

\paragraph{FD} \cite{lessmeier2016condition}: Total 32 bearings are tested under four different operation conditions for rolling bearing fault diagnosis. The motor current signals are recorded for classifying bearing health status, \textit{i.e.}, healthy, artificial damages (D1) and damages from accelerated lifetime tests (D2). We consider each operation condition as an independent domain and select five cross-domain scenarios for evaluation. 


\paragraph{SSC} \cite{goldberger2000physiobank}: Sleep stage classification (SSC) dataset aims to utilize electroencephalography (EEG) signals to identify subject's sleep stage, \textit{i.e.,} wake (W), non-rapid eye movement stage (N1, N2 and N3) and rapid eye movement (REM) stage. Each subject is considered as an independent domain and we select five scenarios for evaluation as previous studies \cite{eldele2021attention}.

\begin{table}[h]
\centering
\begin{adjustbox}{width=0.45\textwidth,center}
\begin{tabular}{c|c|c|c}
\hline
\hline
& {Teacher} & {Student} & Compression Rate \\ \hline
\begin{tabular}[c]{@{}c@{}}No. of Parameters\\ (million)\end{tabular} & 0.2009        & 0.0134     & 14.99 $\times$            \\ \hline
\begin{tabular}[c]{@{}c@{}}No. of FLOPs\\ (million) \end{tabular}                     & 9.328      & 0.661   & 14.11 $\times$             \\ \hline
\hline
\end{tabular}
\end{adjustbox}
\caption{Comparison of model complexity.}
\label{table:model complexity}
\end{table}

\subsection{Experiments Setup}
For our method, a well-trained teacher is a pre-requisite to perform cross-domain knowledge distillation. We adopt \textbf{1D-CNN} as the backbone of our teacher and student models since it consistently outperforms other advanced backbones such as 1D residual network (1D-Resnet) and temporal convolutional neural network (TCN) as indicated in \cite{ragab2022adatime}. We leverage domain-adversarial training of neural networks (DANN) \cite{ganin2016domain} approach to train the teacher. The student is a shallow version of teacher which has less filters. See \textbf{Supplementary} for network details of teacher and student. Table \ref{table:model complexity} summarizes the model complexity of teacher and student in terms of the number of trainable parameters and the number of floating-point operations (FLOPs). We can see that our compact student is about 15$\times$ smaller than its teacher in the aspect of parameters and requires less operations during inference. Furthermore, regarding the evaluation metric, considering the fact that accuracy metric might not be representative for imbalanced dataset, we adopt macro F1-score for all experiments as suggested in \cite{ragab2022adatime}. For all experiments, we repeat 3 times with different random seeds and report the averaged values.

\begin{table*}[h]
\centering
\begin{adjustbox}{width=0.9\textwidth,center}
\begin{tabular}{c||cccccc||cccccc}
\hline
\hline
\multirow{2}{*}{Methods} & \multicolumn{6}{c||}{UCI HAR Transfer Scenario }            & \multicolumn{6}{c}{HHAR Transfer Scenario}                                         \\ \cline{2-13} 
& \multicolumn{1}{c|}{2 $\to$ 11} & \multicolumn{1}{c|}{6$\to$23} & \multicolumn{1}{c|}{7$\to$13} & \multicolumn{1}{c|}{9$\to$18} & \multicolumn{1}{c|}{12$\to$16} & Avg  & \multicolumn{1}{c|}{0$\to$6} & \multicolumn{1}{c|}{1$\to$6} & \multicolumn{1}{c|}{2$\to$7} & \multicolumn{1}{c|}{3$\to$8} & \multicolumn{1}{c|}{4$\to$5} & Avg        \\ \hline
Teacher                       & \multicolumn{1}{c|}{100.00}               & \multicolumn{1}{c|}{96.33}               & \multicolumn{1}{c|}{93.20}               & \multicolumn{1}{c|}{86.45}              & \multicolumn{1}{c|}{68.36}                &  88.87         & \multicolumn{1}{c|}{56.16}              & \multicolumn{1}{c|}{94.10}              & \multicolumn{1}{c|}{54.28}              & \multicolumn{1}{c|}{98.75}              & \multicolumn{1}{c|}{98.67}              &  80.39      \\
Student (src-only)                        & \multicolumn{1}{c|}{60.95}               & \multicolumn{1}{c|}{53.48}               & \multicolumn{1}{c|}{84.61}               & \multicolumn{1}{c|}{35.39}              & \multicolumn{1}{c|}{56.81}                &  58.25         & \multicolumn{1}{c|}{44.45}              & \multicolumn{1}{c|}{66.95}              & \multicolumn{1}{c|}{42.00}              & \multicolumn{1}{c|}{68.84}              & \multicolumn{1}{c|}{65.22}              &  57.49         \\ \hline

DDC                        & \multicolumn{1}{c|}{99.39}               & \multicolumn{1}{c|}{84.44}               & \multicolumn{1}{c|}{84.95}               & \multicolumn{1}{c|}{50.50}              & \multicolumn{1}{c|}{63.29}                & 76.51          & \multicolumn{1}{c|}{52.04}              & \multicolumn{1}{c|}{80.94}              & \multicolumn{1}{c|}{36.78}              & \multicolumn{1}{c|}{74.01}              & \multicolumn{1}{c|}{72.86}              & 63.33          \\
MDDA                       & \multicolumn{1}{c|}{99.44}               & \multicolumn{1}{c|}{93.72}               & \multicolumn{1}{c|}{\textbf{93.20}}      & \multicolumn{1}{c|}{61.20}              & \multicolumn{1}{c|}{59.67}                & 81.45          & \multicolumn{1}{c|}{\textbf{63.51}}     & \multicolumn{1}{c|}{92.44}              & \multicolumn{1}{c|}{51.56}              & \multicolumn{1}{c|}{86.60}              & \multicolumn{1}{c|}{93.01}              & 77.42          \\
HoMM                       & \multicolumn{1}{c|}{\textbf{100.00}}     & \multicolumn{1}{c|}{93.81}               & \multicolumn{1}{c|}{85.23}               & \multicolumn{1}{c|}{68.38}              & \multicolumn{1}{c|}{63.39}                & 82.16          & \multicolumn{1}{c|}{52.48}              & \multicolumn{1}{c|}{90.04}              & \multicolumn{1}{c|}{50.82}              & \multicolumn{1}{c|}{82.18}              & \multicolumn{1}{c|}{91.52}              & 73.41          \\
CoDATS                     & \multicolumn{1}{c|}{88.29}               & \multicolumn{1}{c|}{76.00}               & \multicolumn{1}{c|}{92.23}               & \multicolumn{1}{c|}{70.97}              & \multicolumn{1}{c|}{57.83}                & 77.06          & \multicolumn{1}{c|}{48.96}              & \multicolumn{1}{c|}{92.27}              & \multicolumn{1}{c|}{46.00}              & \multicolumn{1}{c|}{77.91}              & \multicolumn{1}{c|}{75.86}              & 68.20          \\
CDAN                       & \multicolumn{1}{c|}{\textbf{100.00}}     & \multicolumn{1}{c|}{91.43}               & \multicolumn{1}{c|}{91.71}               & \multicolumn{1}{c|}{65.21}              & \multicolumn{1}{c|}{60.44}                & 81.76          & \multicolumn{1}{c|}{45.36}              & \multicolumn{1}{c|}{92.75}              & \multicolumn{1}{c|}{50.79}              & \multicolumn{1}{c|}{91.60}              & \multicolumn{1}{c|}{86.27}              & 73.35          \\
DIRT-T                       & \multicolumn{1}{c|}{\textbf{100.00}}     & \multicolumn{1}{c|}{96.11}               & \multicolumn{1}{c|}{91.32}               & \multicolumn{1}{c|}{67.12}              & \multicolumn{1}{c|}{60.74}                & 83.06          & \multicolumn{1}{c|}{51.48}              & \multicolumn{1}{c|}{93.78}              & \multicolumn{1}{c|}{56.05}              & \multicolumn{1}{c|}{92.51}              & \multicolumn{1}{c|}{97.36}              & 78.24          \\ \hline
JKU                        & \multicolumn{1}{c|}{97.31}               & \multicolumn{1}{c|}{81.54}               & \multicolumn{1}{c|}{91.84}               & \multicolumn{1}{c|}{51.45}              & \multicolumn{1}{c|}{66.44}                & 77.72          & \multicolumn{1}{c|}{49.99}              & \multicolumn{1}{c|}{85.76}              & \multicolumn{1}{c|}{47.65}              & \multicolumn{1}{c|}{84.30}              & \multicolumn{1}{c|}{88.65}              & 71.27          \\
AAD                        & \multicolumn{1}{c|}{92.69}               & \multicolumn{1}{c|}{94.30}               & \multicolumn{1}{c|}{91.52}               & \multicolumn{1}{c|}{72.21}              & \multicolumn{1}{c|}{64.28}                & 83.00          & \multicolumn{1}{c|}{46.01}              & \multicolumn{1}{c|}{93.11}              & \multicolumn{1}{c|}{53.75}              & \multicolumn{1}{c|}{91.03}              & \multicolumn{1}{c|}{92.50}              & 75.28          \\
MobileDA                   & \multicolumn{1}{c|}{88.66}               & \multicolumn{1}{c|}{94.68}               & \multicolumn{1}{c|}{92.83}               & \multicolumn{1}{c|}{75.47}              & \multicolumn{1}{c|}{\textbf{66.67}}       & 83.66          & \multicolumn{1}{c|}{45.17}              & \multicolumn{1}{c|}{93.84}              & \multicolumn{1}{c|}{51.39}              & \multicolumn{1}{c|}{98.39}     & \multicolumn{1}{c|}{78.64}              & 73.49          \\ \hline
\textbf{Proposed}          & \multicolumn{1}{c|}{\textbf{100.00}}               & \multicolumn{1}{c|}{\textbf{96.33}}      & \multicolumn{1}{c|}{\textbf{93.20}}      & \multicolumn{1}{c|}{\textbf{79.77}}     & \multicolumn{1}{c|}{64.91}                & \textbf{86.84} & \multicolumn{1}{c|}{46.66}              & \multicolumn{1}{c|}{\textbf{94.89}}     & \multicolumn{1}{c|}{\textbf{59.20}}     & \multicolumn{1}{c|}{\textbf{98.45}}     & \multicolumn{1}{c|}{\textbf{97.42}}     & \textbf{79.32} \\ \hline
\hline
\end{tabular}
\end{adjustbox}
\caption{Marco F1-score on UCI HAR and HHAR across three independent runs.}
\label{table:uci_har and hhar}
\end{table*}

\begin{table*}[h]
\centering
\begin{adjustbox}{width=0.9\textwidth,center}
\begin{tabular}{c||cccccc||cccccc}
\hline
\hline
\multirow{2}{*}{Methods} & \multicolumn{6}{c||}{FD Transfer Scenario}                                                                                                                                                                                                             & \multicolumn{6}{c}{SSC Transfer Scenario}                                                                                                                                                                                                              \\ \cline{2-13} 
                           & \multicolumn{1}{c|}{0$\to$1} & \multicolumn{1}{c|}{0$\to$3} & \multicolumn{1}{c|}{2$\to$1} & \multicolumn{1}{c|}{1$\to$2} & \multicolumn{1}{c|}{2$\to$3} & Avg       & \multicolumn{1}{c|}{0$\to$11} & \multicolumn{1}{c|}{12$\to$5} & \multicolumn{1}{c|}{16$\to$1} & \multicolumn{1}{c|}{7$\to$18} & \multicolumn{1}{c|}{9$\to$14} & Avg        \\ \hline
Teacher                       & \multicolumn{1}{c|}{84.86}               & \multicolumn{1}{c|}{82.39}               & \multicolumn{1}{c|}{99.59}               & \multicolumn{1}{c|}{90.34}              & \multicolumn{1}{c|}{99.34}                & 91.30          & \multicolumn{1}{c|}{54.20}              & \multicolumn{1}{c|}{66.45}              & \multicolumn{1}{c|}{64.78}              & \multicolumn{1}{c|}{71.48}              & \multicolumn{1}{c|}{72.85}              & 65.95       \\

Student (src-only)                         & \multicolumn{1}{c|}{35.49}               & \multicolumn{1}{c|}{40.46}               & \multicolumn{1}{c|}{87.88}               & \multicolumn{1}{c|}{75.28}              & \multicolumn{1}{c|}{91.13}                & 66.05          & \multicolumn{1}{c|}{33.02}              & \multicolumn{1}{c|}{50.78}              & \multicolumn{1}{c|}{52.25}              & \multicolumn{1}{c|}{57.75}              & \multicolumn{1}{c|}{62.05}              & 51.17       \\ \hline

DDC                        & \multicolumn{1}{c|}{47.31}                & \multicolumn{1}{c|}{59.03}                & \multicolumn{1}{c|}{89.23}                & \multicolumn{1}{c|}{73.57}               & \multicolumn{1}{c|}{89.31}               & 71.69          & \multicolumn{1}{c|}{\textbf{53.09}}      & \multicolumn{1}{c|}{52.29}               & \multicolumn{1}{c|}{57.39}               & \multicolumn{1}{c|}{63.75}               & \multicolumn{1}{c|}{68.53}               & 59.01          \\
MDDA                       & \multicolumn{1}{c|}{71.22}                & \multicolumn{1}{c|}{58.90}                & \multicolumn{1}{c|}{96.06}                & \multicolumn{1}{c|}{84.02}               & \multicolumn{1}{c|}{98.80}               &81.80           & \multicolumn{1}{c|}{32.08}               & \multicolumn{1}{c|}{63.66}               & \multicolumn{1}{c|}{56.98}               & \multicolumn{1}{c|}{65.19}               & \multicolumn{1}{c|}{72.04}               & 57.99          \\
HoMM                       & \multicolumn{1}{c|}{55.44}                & \multicolumn{1}{c|}{48.18}                & \multicolumn{1}{c|}{95.66}                & \multicolumn{1}{c|}{76.39}               & \multicolumn{1}{c|}{96.96}               & 74.53          & \multicolumn{1}{c|}{44.88}               & \multicolumn{1}{c|}{55.47}               & \multicolumn{1}{c|}{56.89}               & \multicolumn{1}{c|}{63.66}               & \multicolumn{1}{c|}{68.87}               & 57.95          \\
CoDATS                     & \multicolumn{1}{c|}{55.72}                & \multicolumn{1}{c|}{64.10}                & \multicolumn{1}{c|}{89.27}                & \multicolumn{1}{c|}{87.58}               & \multicolumn{1}{c|}{91.14}               &77.54           & \multicolumn{1}{c|}{36.69}               & \multicolumn{1}{c|}{61.18}               & \multicolumn{1}{c|}{61.65}               & \multicolumn{1}{c|}{64.47}               & \multicolumn{1}{c|}{62.06}               & 57.21          \\
CDAN                       & \multicolumn{1}{c|}{71.62}                & \multicolumn{1}{c|}{69.53}                & \multicolumn{1}{c|}{97.52}                & \multicolumn{1}{c|}{89.85}               & \multicolumn{1}{c|}{94.48}               & 84.60          & \multicolumn{1}{c|}{33.46}               & \multicolumn{1}{c|}{63.72}               & \multicolumn{1}{c|}{62.04}               & \multicolumn{1}{c|}{65.62}               & \multicolumn{1}{c|}{63.53}               & 57.67          \\
DIRT-T                       & \multicolumn{1}{c|}{76.98}                & \multicolumn{1}{c|}{75.92}                & \multicolumn{1}{c|}{\textbf{99.26}}                & \multicolumn{1}{c|}{\textbf{92.74}}               & \multicolumn{1}{c|}{98.95}      & 88.77         & \multicolumn{1}{c|}{31.71}               & \multicolumn{1}{c|}{\textbf{65.53}}      & \multicolumn{1}{c|}{62.80}               & \multicolumn{1}{c|}{69.87}               & \multicolumn{1}{c|}{69.47}               & 59.88          \\ \hline
JKU                        & \multicolumn{1}{c|}{40.32}                & \multicolumn{1}{c|}{51.44}       & \multicolumn{1}{c|}{88.20}                & \multicolumn{1}{c|}{79.51}               & \multicolumn{1}{c|}{85.80}               & 69.05          & \multicolumn{1}{c|}{41.25}               & \multicolumn{1}{c|}{51.32}               & \multicolumn{1}{c|}{55.34}               & \multicolumn{1}{c|}{66.01}               & \multicolumn{1}{c|}{65.31}               & 55.85          \\
AAD                        & \multicolumn{1}{c|}{64.06}                & \multicolumn{1}{c|}{67.50}                & \multicolumn{1}{c|}{86.94}                & \multicolumn{1}{c|}{79.27}               & \multicolumn{1}{c|}{91.38}               & 77.83          & \multicolumn{1}{c|}{51.40}               & \multicolumn{1}{c|}{58.60}               & \multicolumn{1}{c|}{55.51}               & \multicolumn{1}{c|}{68.77}               & \multicolumn{1}{c|}{59.64}               & 58.78          \\
MobileDA                   & \multicolumn{1}{c|}{40.32}                & \multicolumn{1}{c|}{43.30}                & \multicolumn{1}{c|}{96.82}                & \multicolumn{1}{c|}{76.99}               & \multicolumn{1}{c|}{85.10}               &68.51         & \multicolumn{1}{c|}{53.10}               & \multicolumn{1}{c|}{51.86}               & \multicolumn{1}{c|}{55.60}               & \multicolumn{1}{c|}{65.06}               & \multicolumn{1}{c|}{67.63}               & 58.65          \\ \hline
\textbf{Proposed}          & \multicolumn{1}{c|}{\textbf{78.85}}       & \multicolumn{1}{c|}{\textbf{82.68}}                & \multicolumn{1}{c|}{97.29}       & \multicolumn{1}{c|}{92.14}      & \multicolumn{1}{c|}{\textbf{99.34}}      & \textbf{90.06} & \multicolumn{1}{c|}{44.48}               & \multicolumn{1}{c|}{60.13}               & \multicolumn{1}{c|}{\textbf{62.99}}      & \multicolumn{1}{c|}{\textbf{71.03}}      & \multicolumn{1}{c|}{\textbf{72.21}}      & \textbf{62.17} \\ \hline\hline
\end{tabular}
\end{adjustbox}
\caption{Marco F1-score on Bearing FD and SSC across three independent runs.}
\label{table:wisdm and ssc}
\end{table*}

\begin{figure}[h]
    \centering
    \includegraphics[width=0.45\textwidth]{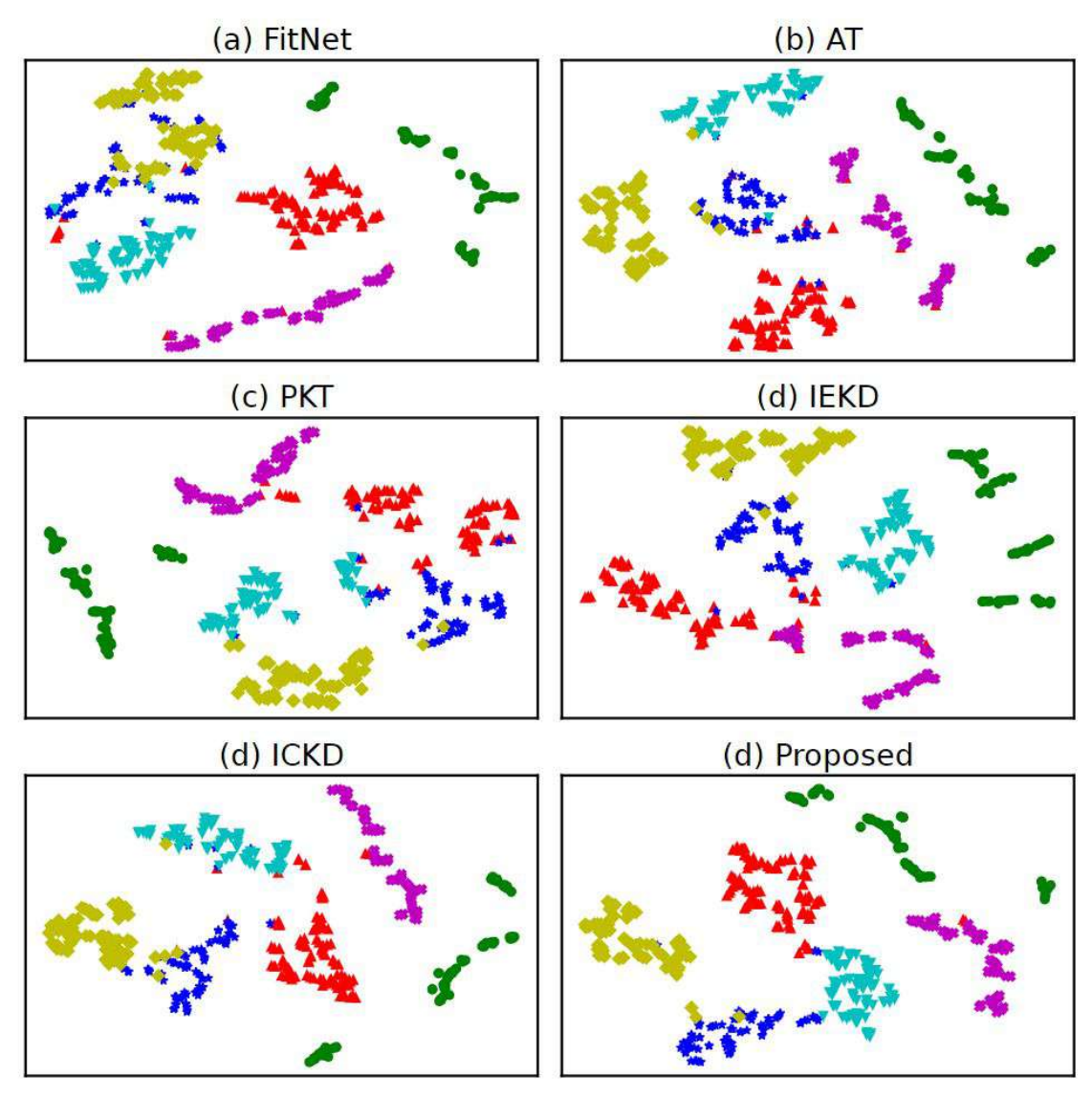}
    \caption{t-SNE of different feature distillation methods on HHAR.}
    \label{fig:figure-t_sne}
\end{figure}

\subsection{Effectiveness of Adversarial Distillation}
Feature-level knowledge from teacher's intermediate layers has already been known as a good extension of logit-based knowledge as DL models are able to learn multiple levels of feature representations \cite{gou2021knowledge}. Various feature-based knowledge distillation approaches have been proposed in existing works. However, for cross-domain KD scenarios, we argue that adversarial learning could more effectively transfer the universal knowledge from teacher to student. To prove it, we first compare the adversarial feature KD with some commonly-used feature distillation approaches: Fitnet \cite{romero2014fitnets}, PKT \cite{passalis2018learning}, AT \cite{zagoruyko2016paying}, IEKD \cite{huang2021revisiting} and ICKD \cite{liu2021exploring}. Please refer to \textbf{Supplementary} for details of each approach. 

It is worth noting that we adapt above methods to our joint logit-level knowledge distillation. The only difference between these methods and ours is the feature distillation part. We utilize the t-SNE to visualize the learnt feature maps of above feature distillation approaches on \textbf{HHAR} dataset. As depicted in Fig. \ref{fig:figure-t_sne}, the features learned from our proposed UNI-KD are more concentrated and all classes are well separated without overlapping. These observations demonstrate that the adversarial feature KD scheme could efficiently transfer the universal knowledge to the student for the cross-domain scenario. More t-SNE visualization results on other three datasets can be found in \textbf{Supplementary}.



\begin{figure*}[h]
\centering
\begin{minipage}{0.5\textwidth}
  \centering
  \includegraphics[width=1.0\textwidth]{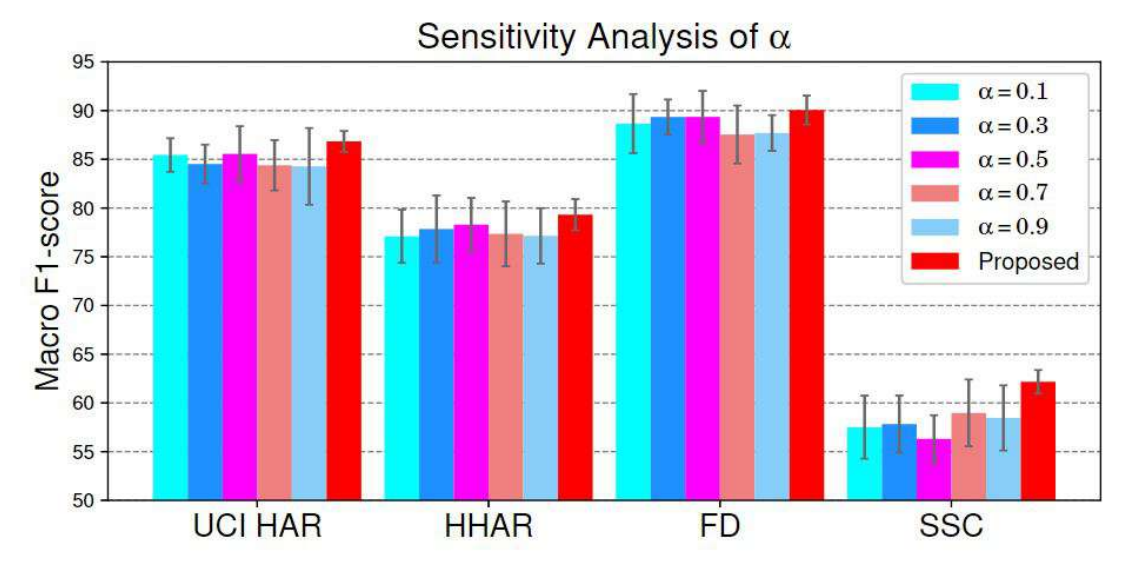}
  \captionof{figure}{Performance with different $\alpha$ values.}
  \label{fig:figure-alpha}
\end{minipage}%
\begin{minipage}{.5\textwidth}
  \centering
  \includegraphics[width=1.0\textwidth]{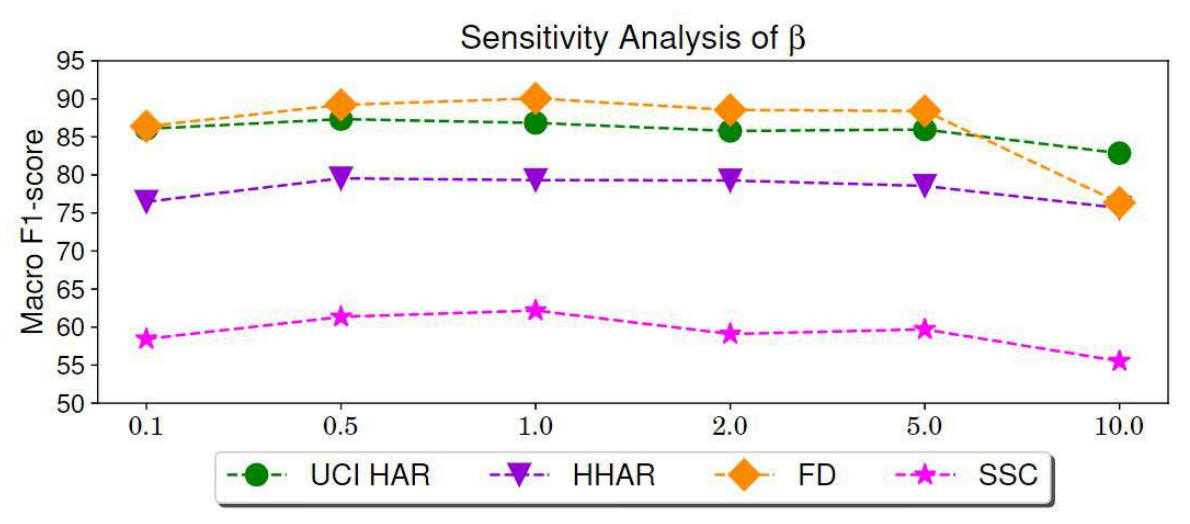}
  \captionof{figure}{Performance with different $\beta$ values.}
  \label{fig:figure-beta}
\end{minipage}
\end{figure*}

\subsection{Benchmark Results and Discussions}
We compare our method with SOTA UDA algorithms, including some discrepancy-based and adversarial learning-based approaches as follows: deep domain confusion (DDC) \cite{tzeng2014deep}, minimum discrepancy for domain adaptation (MDDA) \cite{rahman2020minimum}, higher-order moment matching (HoMM) \cite{chen2020homm}, convolutional deep domain adaptation for time series data (CoDATS)  \cite{wilson2020multi}, conditional adversarial domain adaptation (CDAN) \cite{long2018conditional} and decision-boundary iterative refinement training with a teacher (DIRT-T) \cite{shu2018dirt}. Note that these UDA methods are directly applied to compact student. Meanwhile, we also include the results of some advanced works which integrate UDA with KD as follows: joint knowledge distillation and unsupervised domain adaptation (JKU) \cite{granger2020joint}, adversarial adaptation with distillation (AAD) \cite{ryu2022knowledge} and MobileDA \cite{yang2020mobileda}. See \textbf{Supplementary} for the details of benchmark approaches. Besides, the performance of teacher and student trained on source domain (``Student src-only") are also reported as they can be considered as the upper and lower limits of the compact student. Tables \ref{table:uci_har and hhar} and \ref{table:wisdm and ssc} summarize the evaluation results over different domain adaptation scenarios on four datasets. More experimental results on additional transfer scenarios can be found in \textbf{Supplementary}.

From Tables \ref{table:uci_har and hhar} and \ref{table:wisdm and ssc}, some observations can be found. Firstly, directly applying UDA on a compact student, either the discrepancy-based or adversarial-based UDA approaches, could somehow boost the student's performance on target domain in most of cross-domain scenarios as expected. However, in certain transfer scenarios, negative transfer might also occur. For instance, students trained with the DDC method on 2$\to$7 for \textbf{HHAR} and DIRT-T on 0$\to$11 for \textbf{SSC} even achieve lower performance than the student trained on source only, indicating that those UDA methods might suffer from the inconsistency problem on different domain adaptation tasks. Secondly, the JKU method performs worse than AAD and MobileDA in most of transfer scenarios. The possible reason is that the teacher in JKU is trained together with the student, and it would lead to convergence problem for the student when progressively distilling the knowledge. Moreover, since the teachers used in AAD and MobileDA are only trained on source domain data, the knowledge from these teachers is very limited and biased to source domain, resulting performance degradation. Thirdly, those jointly optimizing KD and UDA methods even under-perform UDA methods like DIRT-T in most transfer scenarios, indicating that improperly transferring teacher's knowledge might decrease student's generalization capability on target data.

Lastly, our method consistently performs the best in terms of averaged macro F1-score over all the four datasets and outperforms other benchmarks on most of transfer scenarios. Moreover, compared with other joint KD and UDA methods, our UNI-KD can significantly reduce the performance gaps between teacher and student with the proposed universal and joint knowledge. This observation demonstrates the effectiveness of our method on the cross-domain model compression scenario. Via the evaluation on various time series domain adaptation tasks, our method can robustly compress the DL models with competitive performance as the complex teacher. 

\subsection{Ablation Study}
There are three key components in our proposed approach: feature-domain discriminator $D_f$, data-domain discriminator $D_d$ and joint knowledge distillation (JKD). To analyze the contribution of each component, we conduct the ablation study as Table \ref{table:ablation} shows. Moreover, to validate the effectiveness of proposed JKD, we also include the standard KD (SKD) in Table \ref{table:ablation}. Some conclusions can be observed from Table \ref{table:ablation} . First, applying universal feature-level KD via integrating $D_f$ upon $D_d$ could consistently improve student's performance over all datasets. However, integrating JKD upon $D_d$ unexpectedly causes performance degradation in \textbf{HHAR} and \textbf{SSC} compared with only employing $D_d$. The possible reason is that logits contain less information than feature maps. Aligning teacher's and student's features could assist the student to learn more general representations than logits. Moreover, our UNI-KD suggests that these two types of knowledge are complementary to each other, and combining them can yield better performance as the last row shows. Furthermore, from the last two rows in Table \ref{table:ablation}, we can conclude that compared with standard KD, our proposed JKD is more effective in the cross-domain scenario. 

\begin{table}[h]
\centering
\begin{adjustbox}{width=0.45\textwidth,center}
\begin{tabular}{cccc|cccc}
\hline
\hline
$D_d$     & $D_f$    & JKD    & SKD        & HAR        & HHAR        & FD          & SSC   \\ \hline
\multicolumn{4}{c|}{Source Only}           & 58.25      & 57.39     & 66.05         & 51.17 \\ \hline
\checkmark    &       &        &           & 82.42      & 76.03     & 83.45         & 60.12 \\
\checkmark    &   \checkmark    & &        & 85.99      & 78.11     & 87.97         & 61.79 \\
\checkmark    &   &  \checkmark      &     & 86.48      & 73.15     & 69.38        & 58.21 \\
\checkmark    & \checkmark & & \checkmark  & 86.31      & 79.01     & 86.29         & 59.68 \\ \hline
\checkmark & \checkmark & \checkmark &     & \textbf{86.84}   & \textbf{79.32} & \textbf{90.06} & \textbf{62.17} \\ \hline
\hline
\end{tabular}
\end{adjustbox}
\caption{Ablation Study for the Proposed UNI-KD.}
\label{table:ablation}
\end{table}

\subsection{Sensitivity Analysis}

There are two hyperparameters (\textit{i.e.}, $\alpha$ and $\beta$) in our proposed approach as shown in Eq. (\ref{eq:equation-loss-total}). For $\alpha$, we propose to gradually increase the importance of JKD loss during the training process as our method relies on an accurate data-domain discriminator. To validate its effectiveness, we compared our adaptive $\alpha$ method with fixed $\alpha$ values as illustrated in Fig. \ref{fig:figure-alpha}. We can see that the proposed adaptive $\alpha$ could consistently achieve better results than fixed $\alpha$. 



For the hyperparameter $\beta$, we utilize the grid search approach to identify the optimal values for different datasets. Fig. \ref{fig:figure-beta} illustrates the performance under different values of $\beta$. We can see that higher value of $\beta$ will result in over-fitting to source data and would decrease the performance as expected. The optimal values for $\beta$ is around $\left[0.5, 1.0\right]$. In all our experiments, we set $\beta= 0.5$ for dataset \textbf{UCI HAR} and \textbf{HHAR} and $\beta= 1.0$ for dataset \textbf{FD} and \textbf{SSC}.

\section{Conclusion}
\label{sec:conclusion}
In this paper, we propose an end-to-end framework for cross-domain knowledge distillation. Our method utilizes an adversarial learning paradigm with a feature-domain discriminator and a data-domain discriminator to improve student's generalization capability on target domain. With our proposed approach, the universal knowledge across both domains and the joint knowledge shared by both domains from a pre-trained teacher can be effectively transferred to a compact student. The experimental results show that the proposed UNI-KD can not only reduce the model complexity but also address domain shift issue. 

\section*{Acknowledgments}

This work is supported by the Agency for Science, Technology and Research (A*STAR) Singapore under its NRF AME Young Individual Research Grant (Grant No. A2084c1067) and A*STAR AME Programmatic
Funds (Grant No. A20H6b0151).


\bibliographystyle{named}
\bibliography{ijcai23}

\end{document}